# YTClickbait21K: Human-Annotated Multimodal Dataset for YouTube Clickbait Detection Across Diverse Channels and Content Categories

Md. Minhazul Islam[1], Md. Tanbeer Jubaer[1], Amith Khandakar[2]*, Shovon Sarker[1], Sumaiya Rahman [1], Md. Masum Mia[2], Mohamed Arselene Ayari[3], Hamed Noori [4]

[1]Department of Computer Science and Engineering, Rajshahi University of Engineering & Technology, Rajshahi, Bangladesh
[2]Department of Electrical Engineering, Qatar University, Doha, Qatar
[3]Department of Civil and Environmental Engineering, Qatar University, Doha, Qatar
[4]SenseNet Inc., Ontario, Canada

*<u>*Corresponding Author:*</u>

**Amith Khandakar**
*Department of Electrical Engineering,*
*Qatar University,*
*Doha, Qatar*
*Email: amitk@qu.edu.qa*

## Abstract

Clickbait content on video-sharing platforms poses a significant challenge to information reliability, yet progress in automated detection has been constrained by the lack of large-scale, high-quality multimodal datasets. We present YTClickbait21K, a human-annotated YouTube clickbait dataset comprising 21,238 videos collected from 40 channels across 29 countries, covering diverse content categories such as news, entertainment, education, and gaming. Each sample includes structured metadata (title, description, engagement statistics) along with associated thumbnail images, enabling comprehensive multimodal analysis. To ensure annotation quality, every video was independently labeled by three annotators using a standardized decision framework that incorporates textual, visual, and cross-modal consistency cues, with final labels determined through majority voting. The dataset exhibits substantial inter-annotator agreement ($\kappa \approx 0.65$), confirming reliable labeling despite the inherent subjectivity of clickbait detection. By combining scale, annotation rigor, and multimodal richness, this dataset provides a robust benchmark for developing and evaluating machine learning models, facilitating research in cross-modal semantic understanding, and advancing automated content moderation systems.

## Background & Summary

Clickbait refers to online content that uses misleading, exaggerated, or emotionally manipulative cues to attract user attention by creating a gap between what is promised and what is actually delivered. Recognized initially in traditional journalism, the phenomenon has grown substantially common and diverse across social media platforms and video-sharing sites[1,2]. Clickbait uses curiosity gaps, hyperbolic language, emotionally charged framing, and forward-referencing expressions to hide key details and encourage interaction[1]. The consequences extend well beyond annoyance: clickbait reduces information quality, affects recommendation algorithms by rewarding engagement over accuracy, and contributes to the spread of misinformation[3]. As the volume and diversity of digital content continue to grow, reliable automated detection has become a major challenge at the intersection of natural language processing, multimedia analysis, and information retrieval. Effective detection at scale requires annotated datasets that accurately represent how clickbait manifests in real-world content, motivating the need for carefully constructed, sufficiently large benchmark resources.

The earliest systematic work treated clickbait as a text classification problem focused on headline linguistics. Chakraborty et al.[1] formalized the problem by showing that stylistic features such as hyperbolic vocabulary, determiners, and stop-word patterns distinguish clickbait from legitimate headlines with 93% accuracy, and also built a real-time browser extension for user-level blocking. Potthast et al.[4] organized the Clickbait Challenge 2017, producing the Webis Clickbait Corpus of 38,517 annotated Twitter posts rated for clickbait intensity on a graded four-point scale by crowdsourced annotators, attracting 13 submitted systems and establishing regression-based clickbait scoring as an important benchmark paradigm. Biyani et al.[3] further demonstrated that article-level informality features in news streams carry a useful discriminative signal, extending the range of features from headline to body text. These foundational studies established text-centric analysis as the dominant paradigm and demonstrated that simple linguistic signals perform well on headline-level datasets.

As research expanded across languages and platforms, a complementary body of work improved text-based datasets with metadata and behavioral signals. Al-Sarem et al.[5] constructed a large-scale Arabic Twitter dataset of 54,893 posts (23,981 clickbait; 30,912 non-clickbait) using semi-supervised pseudo-label expansion and demonstrated that user-based features alongside text effectively support classification. Bronakowski et al.[6] extracted 30 handcrafted semantic and stylistic features from a balanced 32,000-headline English dataset (16,000 per class) and achieved up to 98% accuracy with ensemble classifiers. Liu et al.[7] combined semantic, syntactic, and behavioral features for WeChat content on 9,708 manually annotated samples using a ternary label scheme of non-clickbait, general-clickbait, and malicious-clickbait. For low-resource languages, Muqadas et al.[8] achieved 88% accuracy with a Bi-LSTM model on 1,000 Urdu headlines using sentence-level embeddings, and Nguyen et al.[9] contributed ViClickbait-2025, a curated dataset of 3,414 Vietnamese headlines (1,065 clickbait; 2,349 non-clickbait) annotated by three reviewers with high inter-annotator agreement (Cohen's $\kappa = 0.822$). Elyashar et al.[10] moved toward cross-content analysis by incorporating textual, visual, and behavioral features across Clickbait Challenge datasets of 2,495 and 19,538 Twitter posts, introducing signals that capture discrepancies between a social post and its linked article. Despite this enrichment, datasets in this line of work still do not model thumbnail or video-frame visual content, leaving the cues most typical of deceptive content on video platforms unexplored.

The recognition that clickbait on video platforms is inherently visual as well as textual prompted a shift toward multimodal approaches. Gamage et al.[11] proposed BaitRadar, a multimodal deep learning framework integrating title, thumbnail, comments, tags, statistics, and audio transcript, achieving high accuracy on a dataset of approximately 14,000 videos. However, the study relies on channel-level annotation labels inherited from prior work, Zannettou et al.[12] , rather than explicit video-level annotations, which introduces potential label noise and bias toward channel-specific patterns. Vitadhani et al.[13] focused the focus to thumbnail cues by applying OCR and facial recognition to 250 manually annotated videos, confirming that embedded thumbnail text and facial imagery carry useful information. Varshney and Vishwakarma[14] introduced the MVD dataset comprising 987 YouTube videos (474 clickbait; 513 non-clickbait), emphasizing speech-title similarity and audience reactions as proxies for deception. Mowar et al.[15] contributed BollyBAIT, 1,000 manually labeled Bollywood-focused YouTube videos that explicitly consider the relationship between title, thumbnail, and actual video content. Sung et al.[16] extended the multimodal perspective to 2,247 Facebook video posts in the VMH dataset, confirming that unimodal models perform substantially worse than multimodal approaches and providing direct evidence that cross-modal semantic incongruence is the main mechanism of deception. Rahman et al.[17] further found that multimodal fusion outperforms single-modality baselines for YouTube clickbait, though limited annotation transparency reduces reproducibility. Complementing automated efforts, Al-Ali and Hamzeh[18] identified through qualitative analysis of 100 Arabic YouTube thumbnails that emotional imagery, emojis, and curiosity-inducing visual-textual complementarity are key elements of thumbnail-based clickbait.

Efforts to construct large-scale multimodal benchmarks have exposed a persistent tension between scale and annotation quality. Imran et al.[19]introduced BaitBuster-Bangla, comprising over 253,000 YouTube videos from 58 Bangla channels with thumbnails, textual metadata, and engagement features. However, only approximately 10,000 samples are manually annotated; the remainder rely on automated or AI-generated labels, making the labels' reliability difficult to verify. Despite a multimodal design, the baseline modeling remains largely text-focused and does not fully exploit visual signals. Naveed et al.[20] proposed ThumbnailTruth, a cross-cultural dataset of 2,843 YouTube videos from eight countries (1,359 misleading; 1,484 non-misleading), combining thumbnails, subtitle transcripts, and LLM-generated video descriptions to study thumbnail-content mismatch directly. This is one of the most important recent efforts given its focus on thumbnail misleadingness and cross-cultural diversity, but the dataset remains relatively small and depends on LLM-generated descriptions rather than raw visual content. In adjacent work on video-level misinformation, Kim et al.[21] and Zhang et al.[22] demonstrated that hybrid and graph-based cross-modal reasoning frameworks yield strong detection improvements, reinforcing the broader principle that video deception cannot be adequately characterized through text alone.

Taken together, the literature traces a clear progression from text-only headline classification to metadata-improved datasets, and then to genuinely multimodal approaches that incorporate thumbnails, transcripts, and engagement signals. Yet persistent limitations appear at each stage: many datasets are text-centric or restricted to a single language or domain; [23,19,15] multimodal datasets frequently lack explicit modeling of the semantic relationship between thumbnail, title, and video content; annotation quality is inconsistently reported; and even the largest-scale resources depend substantially on automated labels. **Table 1** summarizes existing datasets and highlights these gaps. Manually annotated multimodal datasets with content diversity remain small relative to the requirements of modern deep learning. Goodfellow et al.[24] recommend at least 5,000 labeled examples per category for supervised deep learning to achieve acceptable performance. With 21238 manually annotated samples, the present dataset substantially exceeds this threshold and is designed to support reliable training and evaluation of multimodal architectures. Unlike prior work that relies on inherited channel-level labels, automated labeling, or narrow domain samples, the proposed dataset is built through a careful multi-annotator pipeline with explicit quality control and administrative verification at the instance level, ensuring that labels accurately reflect video-level deceptive intent across diverse content categories.

***Table 1. Comparison of existing clickbait datasets across platforms, modalities, annotation strategies, and key limitations***

| Ref. / Dataset | Year | Platform | Total | Clickbait | Non-CB | Modalities | Annotation | Key Limitation |
|---|---|---|---|---|---|---|---|---|
| Chakraborty et al. (Webis CB 2016)[1] | 2016 | News/Web | ~15,000 | ~7,500 | ~7,500 | Text (headline) | Domain-based crawl | Text only; no visual |
| Potthast et al. (Webis CB 2017)[4] | 2018 | Twitter | 38,517 | ~4,761 | ~33,756 | Text (tweet + article) | Crowdsourced (AMT) | Text only; graded labels |
| Al-Sarem et al. (Arabic CB)[5] | 2021 | Twitter | 54,893 | 23,981 | 30,912 | Text + user metadata | Semi-supervised (PLL) | Arabic only; no visual |
| Bronakowski et al. (Semantic CB)[6] | 2023 | News/Web | 32,000 | 16,000 | 16,000 | Text (headline) | Public dataset; manual | Text only; English only |
| Liu et al. (WeChat CB)[7] | 2022 | WeChat | 9,708 | N/R | N/R | Text (title) | Manual (3 annotators) | Ternary labels; no visual |
| Zheng et al. (CB Challenge+CN)[23] | 2020 | News/Web | ~24,000 | N/R | N/R | Text (headline + body) | Manual + crowdsourced | Limited to text modality |
| Muqadas et al. (Urdu CB) [8] | 2025 | News | 1,000 | N/R | N/R | Text (headline) | Manual (experts) | Urdu only; very small scale |
| Nguyen et al. (ViClickbait-2025)[9] | 2025 | News | 3,414 | 1,065 | 2,349 | Text (headline) | Manual (κ = 0.822) | Text only; no visual |
| Elyashar et al. (CB Challenge)[10] | 2022 | Twitter | 19,538 + 2,495 | N/R | N/R | Text + visual + behavioral | CB Challenge labels | Handcrafted features only |
| Gamage et al. (BaitRadar)[11] | 2021 | YouTube | 1,4000 | 8,591 | 5,049 | Title + thumbnail + tags + transcript + comments + stats | Channel-level assumption | Small scale; weak label source |
| Vitadhani et al.(Thumbnail CB)[13] | 2021 | YouTube | 250 | N/R | N/R | Thumbnail (OCR + face) + title | Manual | Very small; limited diversity |
| Varshney & Vishwakarma (MVD)[14] | 2021 | YouTube | 987 | 474 | 513 | Title + transcript + comments + channel | Manual (per-video) | No direct thumbnail modeling |
| Mowar et al. (BollyBAIT)[15] | 2021 | YouTube | 1,000 | N/R | N/R | Title + thumbnail + video content | Manual | Small; Bollywood domain only |
| Sung et al. (VMH)[16] | 2023 | Facebook | 2,247 | N/R | N/R | Video + headline + transcript + metadata | Manual + rationales | Facebook only; limited scale |
| Rahman et al. (YT Deceptive CB)[17] | 2023 | YouTube | N/R | N/R | N/R | Title + thumbnail + metadata | Manual | Annotation details unclear |
| Imran et al. (BaitBuster-Bangla) [19] | 2024 | YouTube | 253,070 | N/R | N/R | Thumbnail + text + engagement | ~10K manual; rest auto/AI | Weak labels at scale; text-focused baseline |
| Naveed et al. (ThumbnailTruth)[20] | 2025 | YouTube | 2,843 | 1,359 | 1,484 | Thumbnail + transcript + LLM description | Manual (8 countries) | Small; LLM-generated descriptions |
| **YTClickbait21K (Ours)** | **2025** | **YouTube** | **21,238** | **9,653** | **11,585** | **Thumbnail (local JPG/WebP) + title + metadata; video & audio via URL** | **Manual; multi-annotator; admin QC** | — |

# Methods

The construction of the YTClickbait21K dataset followed a structured three-phase pipeline: (1) automated collection of YouTube video metadata via the official YouTube Data API v3 using a purpose-built FastAPI-based framework, (2) systematic human annotation through a custom web-based labeling platform, and (3) data aggregation and quality verification prior to final export. **Figure 1** provides a schematic overview of the complete workflow.

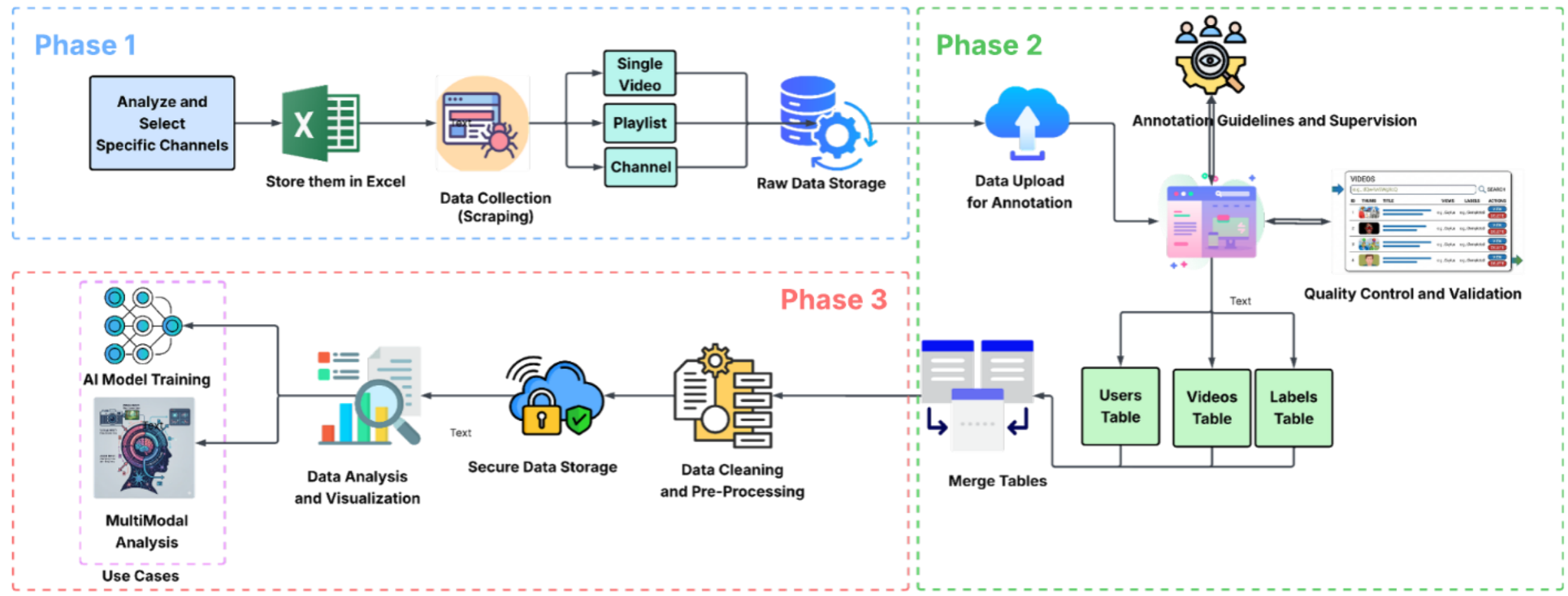

***Figure 1. Overview of the dataset construction pipeline, including data collection, annotation, and quality validation stages***

# Data Collection

Data collection is a critical component of this study, as the quality, reliability, and ethical compliance of the dataset directly influence the validity of the subsequent analysis. To ensure a scalable and policy-compliant approach, this study leverages YouTube as the sole data source, given its global reach, wide range of content categories, and the frequent presence of clickbait practices reported in prior studies on video-sharing platforms.

This YouTube data collection project adheres to YouTube's Terms of Service by using the official YouTube Data API v3, which is Google's authorized framework for accessing YouTube metadata. Data collection was carried out using a custom web application built with FastAPI, in which authenticated API requests retrieved structured video metadata. This setup allowed efficient and reliable retrieval of metadata directly from YouTube's official services without downloading the actual video content, ensuring both speed and data completeness. All data were collected strictly from publicly available sources, and no personal or private user information was accessed or stored. Furthermore, no copyrighted video or media content was downloaded or retained; only metadata and publicly accessible resource links were collected in full compliance with YouTube's terms of service and Google's data usage policies.

The collection framework was designed with three distinct methods. (1) Individual video URL ingestion processes a single video via the API and returns its metadata as one record. (2) Playlist-level extraction retrieves metadata for all videos within a playlist in a structured manner using API endpoints. (3) Channel-level collection gathers videos from a channel by accessing its publicly available video resources through the API, with retrieval limited by predefined quotas. All three methods use the same metadata extraction pipeline, which ensures consistency across the collected data. The YouTube Data API v3 provides a free usage tier with a daily quota limit, which was sufficient for this research-scale data collection. During the collection process, entries that could not be retrieved properly due to API errors, quota limits, or missing information were excluded to maintain the quality of the final dataset. Throughout the collection process, strict adherence to YouTube's usage policies was maintained, and only information made publicly accessible by content creators was included.

For each video, the metadata extraction process collected a set of structured information fields. Each video was assigned a unique identifier, and its corresponding URL was generated automatically. The dataset also includes a thumbnail link that points to the highest-resolution image available for the video. Textual information, such as the video title and description, was also recorded; if a title was unavailable, the identifier was used as a fallback, while the description was limited to a fixed length to avoid overly large entries. In addition, information about the content creator was included, such as the channel name and its unique identifier. No personally identifiable information beyond publicly available channel-level identifiers was collected or inferred.

Basic engagement and content-related details were also captured, including the number of views, the number of likes when available, the video duration in seconds, and the upload date in a standard format. To support multimodal analysis, thumbnail images were optionally retrieved using their publicly available URLs and stored locally in standard image formats (JPG or WebP) for offline processing. This enables integration of visual features alongside textual metadata in downstream machine learning tasks. Importantly, only publicly accessible thumbnail resources were utilized, and no video or audio content was downloaded or retained. Together, these fields provide sufficient context to support clickbait detection, multimodal learning, and further analysis tasks.

# Channel Selection and Deduplication

A total of 40 YouTube channels were selected to form the dataset, covering various content categories such as news media, documentary production, entertainment, gaming, and educational content. Channel selection was preceded by a preliminary study in which a broad pool of candidate channels was examined with respect to subscriber count, average view engagement, and content category. Channels were ultimately chosen on the basis of two complementary criteria: popularity, to ensure that the selected channels reflect the kind of content that real audiences engage with at scale, and content divergence, to maximize heterogeneity in publishing norms, audience expectations, and clickbait strategies across the final selection. This deliberate combination of high-reach and topically diverse channels ensures that the dataset is neither biased toward any single content niche nor dominated by low-visibility content unlikely to appear in real-world recommendation contexts. The selected channels represent content from multiple countries and regions, ensuring that the dataset captures global variations in content style and clickbait practices rather than being limited to a single geographical context. **Figure 2** shows the geographic distribution of the 40 selected channels across 29 countries.

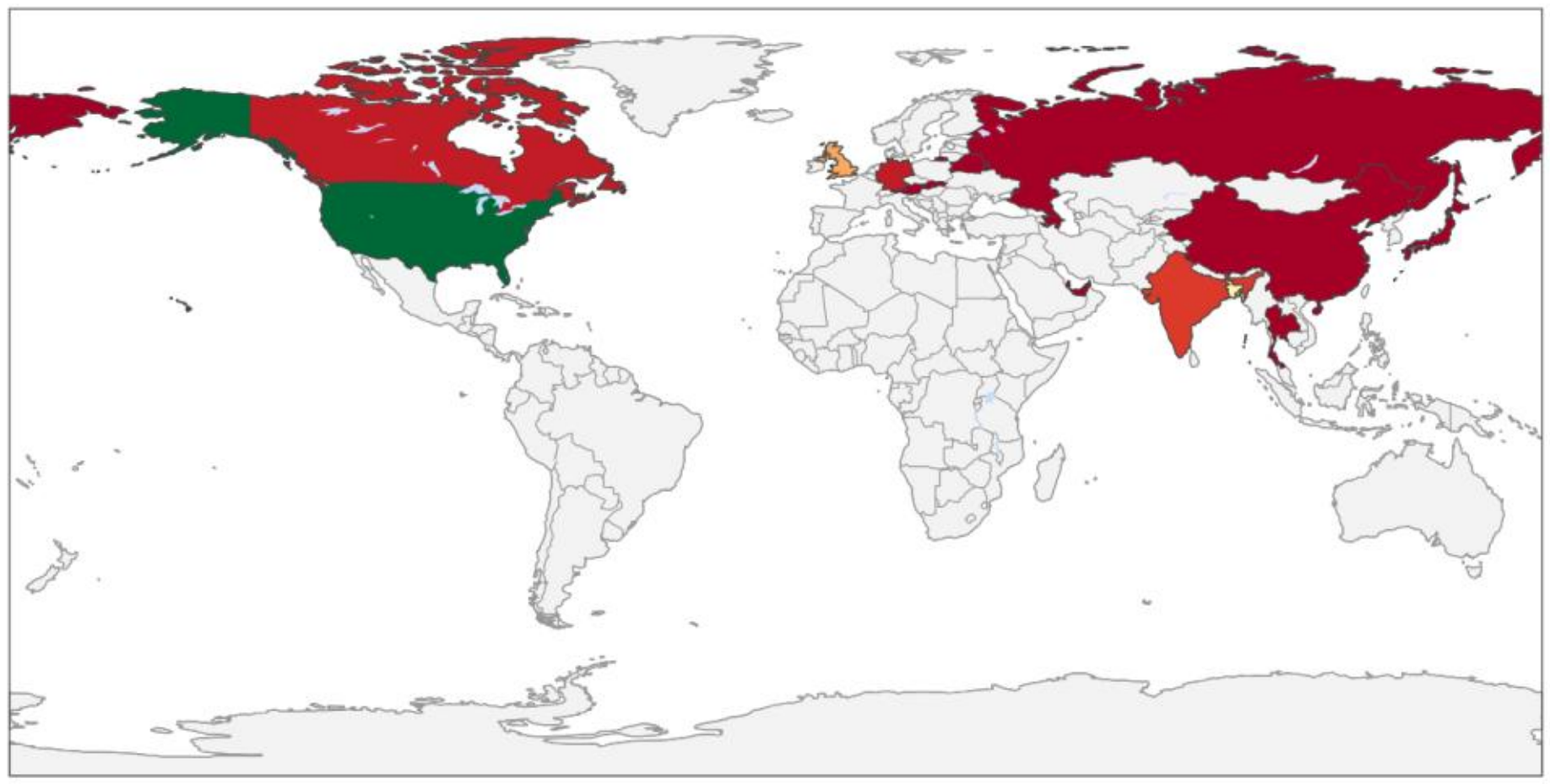

***Figure 2. Geographic distribution of selected YouTube channels across 29 countries.***

To avoid duplicate records, the system used a first-occurrence-wins deduplication method based on video_id within each scraping session. All collected data were stored in an in-memory session during collection and exported as comma-separated value files after each scraping batch. A persistent scraper log tracked the datetime, source URL, scraping mode, channel name, and number of videos added for each operation, ensuring full accountability of the collection process.

# Annotation Platform

To enable systematic labeling at scale, a dedicated web-based annotation platform was created specifically for this project. The platform was built with PHP and a MySQL-based backend, enabling reliable data handling and smooth user interactions. The website is hosted online and can be accessed at https://clickbait.rf.gd, making it easily accessible to annotators from different locations. This architecture supports deployment on standard hosting services, offering operational flexibility.

The platform features role-based access control that differentiates between standard annotators and administrators. Administrators have exclusive access to functions such as importing and exporting data, managing annotator accounts, monitoring overall annotation progress, and distributing annotation guidelines.

The system maintains structured storage of user information, video data, and annotation records. It ensures that each video receives multiple independent annotations while preventing the same annotator from labeling the same video more than once. The platform also tracks annotation progress for each video and marks entries as complete once the required number of annotations is reached. Each annotation is recorded with the annotator's identity, assigned label, confidence level, and a system-generated timestamp, ensuring traceability and consistency throughout the labeling process.

# Annotation Interface and Workflow

Producing reliable labels for a dataset of this scale required more than a well-designed interface; it required annotators who genuinely understood the nuance and subjectivity involved in identifying clickbait. Before any annotation began, the research team developed a detailed annotation guideline document that defined clickbait, described the five decision criteria used for labeling, and provided worked examples of clear-cut and ambiguous cases drawn from diverse content categories. This document was distributed to all annotators in advance of their first session. In addition, a series of synchronization meetings was held with the full annotator group to discuss the guidelines, resolve disagreements over example cases, calibrate a shared understanding of edge cases, and answer questions about content categories that did not fit neatly into the binary label scheme. These sessions were not one-time events; the team maintained ongoing communication with annotators throughout the data collection period, issuing clarification notes when recurring sources of confusion were identified and holding supplementary briefings as new content categories entered the annotation queue.

To further support consistent judgment, annotators were provided with a structured decision framework that organized the key cues they should consider when evaluating a video. This framework was presented as a set of five interconnected criteria covering title analysis, thumbnail inspection, cross-modal consistency, engagement context, and an overall faithfulness judgment. It was designed not as a mechanical checklist but as an ordered guide to human reasoning: annotators were expected to weigh these criteria holistically, exercise judgment, and assign a single binary label indicating whether the video was designed to mislead viewers. The framework is summarized in **Table 2.**

***Table 2: Annotation decision framework outlining criteria used for clickbait labeling***

| **Decision Framework for Clickbait Annotation** | |
|---|---|
| Criterion 1: Title analysis | Does the title withhold key information to induce curiosity, use excessive capitalization, or employ hyperbolic or sensational language that does not reflect the video content? |
| Criterion 2: Thumbnail inspection | Does the thumbnail use misleading imagery, appear digitally manipulated, or present visual elements that do not correspond to the actual content of the video? |

| Criterion 3: Cross-modal consistency | Is there a meaningful disconnect between the title, the description, and the actual video content? Does the content fail to deliver what the title or thumbnail implies? |
|---|---|
| Criterion 4: Engagement context | Do engagement indicators (view count, like ratio) or channel history suggest a pattern of deceptive presentation rather than genuine content? |
| Criterion 5: Overall faithfulness | Taken as a whole, does the video honestly and accurately represent its content to a viewer encountering it for the first time? If clearly faithful, assign Non-Clickbait; if ambiguous but suspicious across multiple criteria, assign Clickbait. |

Annotators accessed a dedicated web-based dashboard after authentication, where they could view the full queue of videos pending annotation. The interface supported sequential navigation through unannotated items and manual selection of specific videos from the full list. For each video, the interface displayed the title, thumbnail, description, channel name, view count, and like count. Labels assigned by other annotators were deliberately concealed throughout the session to prevent information cascade effects and preserve independence in labeling decisions**. Figure 3** provides a screenshot of the annotation platform.

Each annotation consisted of two components. Annotators first assigned a binary label, with 1 indicating clickbait and 0 indicating non-clickbait. They then recorded a confidence score on a 1 to 5 ordinal scale, with 1 representing very low certainty and 5 representing very high certainty. This dual-component design was deliberate: capturing the confidence score alongside the binary decision allows downstream analyses to identify genuinely ambiguous instances, apply confidence-weighted aggregation when resolving disagreements, and exclude or separately analyze low-confidence cases during quality control. Annotators were encouraged to use the full range of the confidence scale honestly rather than defaulting to high scores, and this expectation was reinforced during the synchronization meetings.

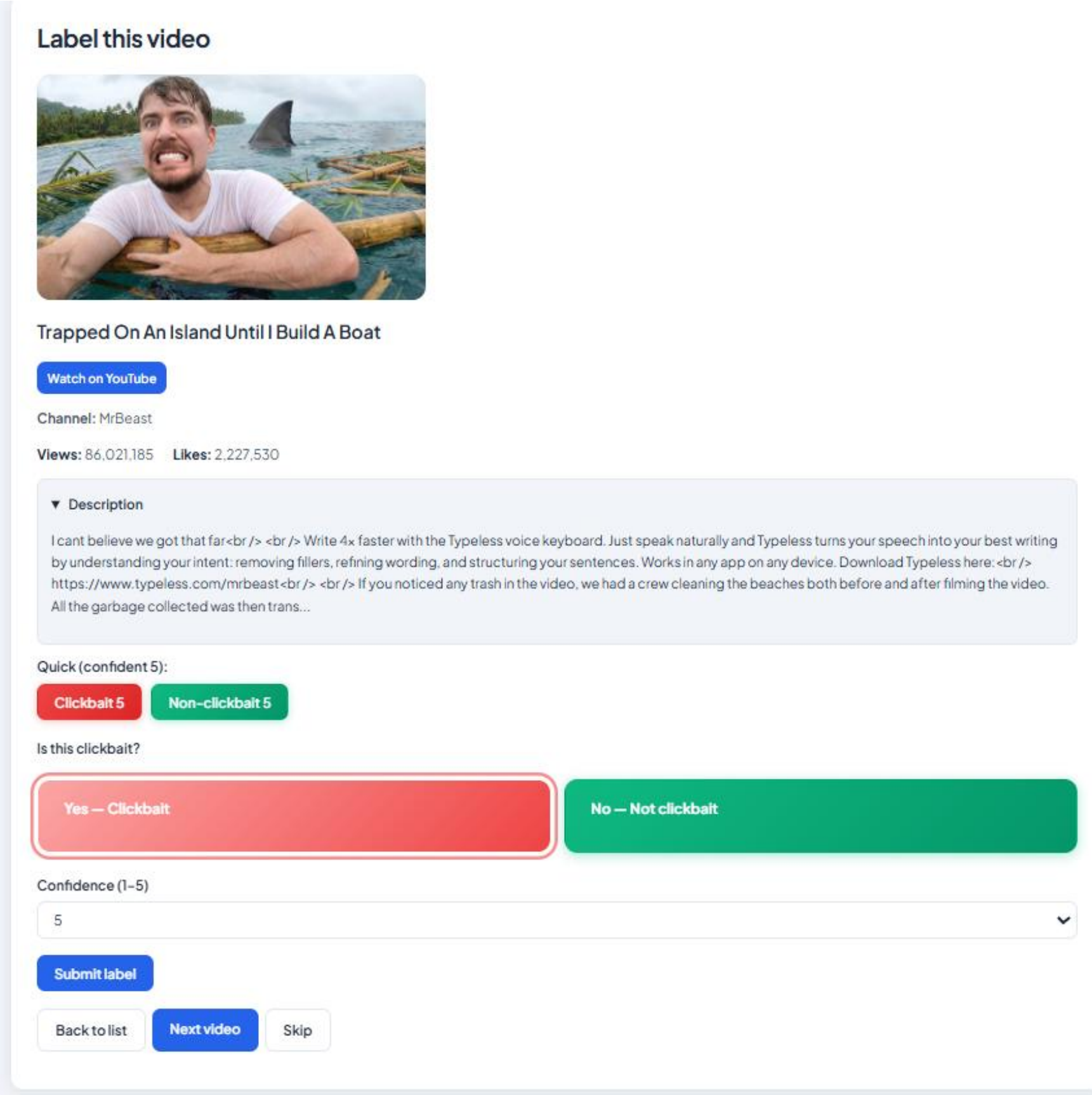

***Figure 3. Web-based annotation interface used for labeling clickbait and non-clickbait videos.***

To further ensure label reliability, an administrative review layer was incorporated into the workflow. A designated reviewer examined annotations flagged as low-confidence or as potential disagreements between annotators, resolved outstanding cases through re-review or arbitration, and monitored annotation patterns across individuals and sessions for signs of drift or systematic bias. This layer of oversight was particularly important for ambiguous videos, where the five criteria pointed in conflicting directions, and it ensured that the final labels reflected a consistent, well-reasoned standard rather than the idiosyncratic judgments of individual annotators.

Annotation guidelines were provided as PDF documents uploaded by administrators and remained accessible throughout the process. These guidelines defined the criteria for identifying clickbait while preserving procedural independence. The overall annotation workflow, from video ingestion to independent labeling and validation, is summarized in **Figure 4**.

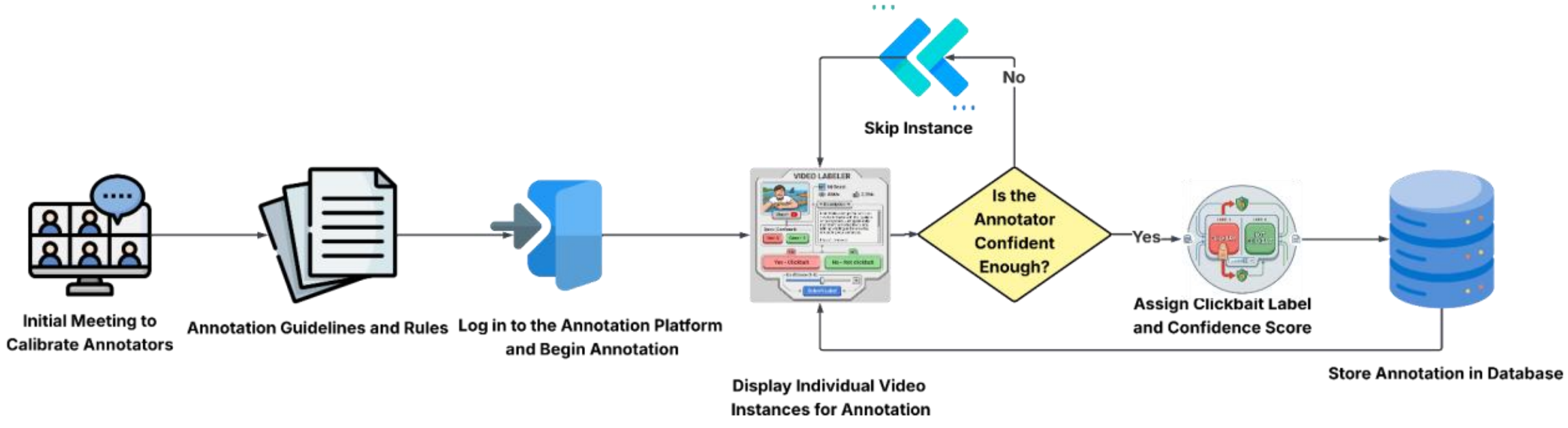

***Figure 4. End-to-end annotation workflow from guideline preparation to final label assignment and storage.***

# Annotation Protocol and Quality Control

A total of 15 annotators participated in labeling the complete dataset. All annotators were at minimum university students with demonstrated knowledge of online content and digital media, and each was individually briefed on the annotation guidelines and decision framework prior to contributing any labels. From this pool, each video was assigned to exactly three independent annotators. Annotators submitted labels independently, with no access to other raters' decisions or confidence scores at any point during the annotation process.

The system enforced a strict one-label-per-annotator-per-video constraint, preserving independence across all annotations. Every video in the final dataset received the required three evaluations, and this multi-annotator design strengthens reliability by reducing the influence of individual biases or errors.

# Final Label Determination

The final label for each video was determined through majority voting across the three independent annotations. Under this scheme, a video receives a final label of 1 (clickbait) if at least two of the three annotators assigned that classification, and a final label of 0 (non-clickbait) otherwise.

This approach is robust to individual annotator errors and is consistent with established practices in crowdsourced annotation research. The resulting label distribution reflects a moderately balanced dataset, ensuring suitability for downstream machine learning tasks without severe class imbalance.

# Inter-Annotator Agreement

Inter-annotator agreement was evaluated using pairwise Cohen's Kappa coefficients, which account for chance agreement and are widely regarded as a standard reliability measure for binary annotation tasks. The computed Kappa scores were $\kappa = 0.6634$ between Annotators 1 and 2, $\kappa = 0.6494$ between Annotators 2 and 3, and $\kappa = 0.6401$ between Annotators 1 and 3. The mean pairwise Cohen's Kappa across all annotator pairs was 0.6510, indicating substantial agreement according to the Landis and Koch scale (0.61–0.80). These results demonstrate strong consistency among annotators and confirm the overall reliability and robustness of the labeling process.

To further analyze agreement patterns, the dataset was examined at the instance level. A large majority of videos exhibited full agreement among all three annotators, while a smaller portion showed partial disagreement. Such discrepancies are expected in clickbait detection tasks, as the distinction between clickbait and non-clickbait is often subtle, context-dependent, and influenced by subjective interpretation of linguistic cues and user expectations. Prior work, [25], highlights that YouTube clickbait detection is challenging, that titles alone are not reliable, and that the authors needed multiple pieces of evidence, such as video content, comments, and user profiles, to detect clickbait on YouTube. It also explicitly notes that relying solely on some cues can fail, leading to variability even among human annotators.

Additionally, annotators reported consistently high confidence in their labeling decisions, indicating that disagreements were not due to uncertainty but rather to genuine differences in interpretation. This further supports the idea that clickbait detection is a complex task rather than a simple binary classification problem, as also reflected in the inter-annotator agreement summarized in **Table 3**.

***Table 3. Summary of inter-annotator agreement measured using Cohen's Kappa across annotator pairs***

| Annotator Pair | Cohen's Kappa (κ) | Agreement Level |
|---|---|---|
| Annotator 1 vs 2 | 0.6634 | Substantial |
| Annotator 2 vs 3 | 0.6494 | Substantial |
| Annotator 1 vs 3 | 0.6401 | Substantial |
| Average | 0.6510 | Substantial |

Overall, the substantial agreement, combined with controlled multi-annotator design and high annotation confidence, indicates that the dataset provides a reliable foundation for downstream clickbait detection tasks while appropriately capturing the inherent subjectivity of the problem.

# Confidence Score Analysis

The average confidence score across all annotation records was 4.79 out of 5.00, indicating that annotators consistently expressed a high degree of certainty in their labeling decisions. A substantial majority of annotations were assigned high-confidence values ($\geq 4.5$), indicating that most labeling decisions were made with strong conviction.

The confidence scores are preserved as an auxiliary field in the final dataset, enabling more flexible and robust downstream analysis. In particular, they support confidence-weighted aggregation strategies, low-confidence instance filtering, and sensitivity analyses to assess model robustness across varying levels of annotation certainty. This additional layer of information enhances the dataset's practical utility for both research and real-world applications.

# Data Records

The dataset associated with this work consists of a single primary data file, Final_data.csv, deposited in the Figshare public data repository and accessible without restriction. Additionally, a separate folder containing thumbnail images is included, totaling approximately 2.41 GB. The file consolidates YouTube video metadata with the complete multi-annotator labeling records and the derived consensus label for each video into a unified, analysis-ready structure. The organization of the dataset, including the folder structure and file placement, is illustrated in **Figure 5**. The following subsections describe the file format, column schema, distributional properties, and content characteristics of this record in detail.

## Files Structure

```
Final_data.csv
Thumbnail/
  ├── {video_id}.jpg
  └── {video_id}.webp
```

***Figure 5. Structure of the dataset directory, including metadata file and associated thumbnail images***

# Primary Data File

The primary data record is *Final_data.csv*, a UTF-8-encoded comma-separated values file containing 21,238 rows and 26 columns. Each row represents one unique YouTube video, identified by its video_id field, which serves as the primary key across the dataset. The file size is approximately 33.25 MB. No special software is required to read this file; it is compatible with standard data analysis environments, including Python (pandas), R, and spreadsheet applications that support UTF-8 encoded CSV files with quoted fields, which are necessary to accommodate video titles and descriptions containing commas and other special characters. **Table 4** provides the complete column schema.

**Table 4. Schema of the main dataset (Final_data.csv) detailing column names, data types, and descriptions**

| Column Name | Data Type | Description |
|---|---|---|
| video_id | String | Unique YouTube video identifier. Functions as the primary key; no duplicate values are present in the dataset. |
| url | String (URL) | Full YouTube watch URL derived from the video identifier, in the format https://www.youtube.com/watch?v={video_id}. |
| thumbnail | String (URL) | Direct URL of the highest-resolution thumbnail available for the video at the time of collection. |
| title | String | Full video title as displayed on YouTube. This field constitutes the primary textual feature for clickbait classification tasks. |
| description | String | Video description truncated to a maximum of 5,000 characters. |
| channel_name | String | Display name of the YouTube channel that published the video |
| channel_id | String | Alphanumeric unique identifier of the publishing channel, as assigned by YouTube. |
| views | Integer | Total view count recorded at the time of data collection. Ranges from 5 to 905,033,299 across the dataset. |
| likes | Integer | Total like count recorded at the time of data collection. |
| duration | String | Video duration expressed in seconds with a trailing 's' suffix ( '420s'). Ranges from 9 seconds to 87,536 seconds. |
| upload_date | Date (YYYY-MM-DD) | Date on which the video was published to YouTube, normalized from the native YYYYMMDD format. |
| created_at | Timestamp | Date and time at which the video record was inserted into the annotation platform database. |

| num_labels | Integer {1–3} | Total number of annotation labels received by the video. All records in the final dataset carry a value of 3, confirming complete annotation coverage. |
|---|---|---|
| label_agreement_count | Integer {1–3} | Number of annotators whose label matches the majority-voted final label. A value of 3 indicates unanimous agreement; a value of 2 indicates majority agreement with one dissent. |
| annotator_1 | String | Username of the first annotator assigned to the video. |
| annotation_label_1 | Binary {0, 1} | Binary clickbait classification assigned by Annotator 1. A value of 1 indicates clickbait; a value of 0 indicates non-clickbait. |
| annotation_confidence_1 | Integer [1–5] | Ordinal confidence score submitted by Annotator 1 alongside the binary label. A score of 1 denotes very low certainty; a score of 5 denotes very high certainty. |
| annotator_2 | String | Username of the second annotator assigned to the video. |
| annotation_label_2 | Binary {0, 1} | Binary clickbait classification assigned by Annotator 2. |
| annotation_confidence_2 | Integer [1–5] | Ordinal confidence score submitted by Annotator 2. |
| annotator_3 | String | Username of the third annotator assigned to the video. |
| annotation_label_3 | Binary {0, 1} | Binary clickbait classification assigned by Annotator 3. |
| annotation_confidence_3 | Integer [1–5] | Ordinal confidence score submitted by Annotator 3. |
| final_label | Binary {0, 1} | Consensus label derived through majority voting across the three independent annotations. A video is assigned a value of 1 if at least two of the three annotators classified it as clickbait, and 0 otherwise. |
| average_confidence | Integer [1–5] | Mean confidence score computed across all available annotator confidence values for the video, providing a continuous measure of collective annotation certainty. |
| local_thumbnail | String (path) | Local file path reference for the downloaded thumbnail image. Files are stored in JPG or WebP format depending on availability, following the naming convention Thumbnail\videoid.jpg or Thumbnail\videoid.webp. These local copies are provided for offline multimodal experiments. Video content and audio transcriptions are not bundled with the dataset due to storage constraints but can be retrieved via the YouTube URLs provided for each record. |

# Label Distribution

The final label distribution across the 21,238 records reflects a moderately imbalanced binary classification dataset. A total of 9,653 videos (45.45%) were assigned a final label of 1 (clickbait) and 11,585 videos (54.55%) a final label of 0 (non-clickbait) through majority voting. This distribution reduces the risk of severe class imbalance artifacts in classifier performance metrics, making the dataset suitable for direct use without aggressive resampling. **Table 5** summarizes the label distribution.

*Table 5. Final label distribution of clickbait and non-clickbait samples in the dataset*

| Label | Count | Percentage | Interpretation |
|---|---|---|---|
| Clickbait (1) | 9,653 | 45.45% | Videos classified as clickbait by the majority vote of three independent annotators. |
| Non-Clickbait (0) | 11,585 | 54.55% | Videos classified as non-clickbait by the majority vote of three independent annotators. |
| Total | 21,238 | 100% | Complete dataset after deduplication and annotation. |

# Content Characteristics

The dataset spans 40 YouTube channels drawn from a range of content categories, including news media, documentary production, entertainment, gaming, and educational content. **Table 6** summarizes the distributional characteristics of the primary content metrics.

*Table 6. Statistical distribution of key content metrics, including views, likes, and video duration*

| Metric | Mean | Median | Minimum | Maximum |
|---|---|---|---|---|
| Views per video | 6,057,442 | 344,256 | 8 | 905,033,299 |
| Likes per video | 149,677 | 5,737 | 1 | 30,196,410 |
| Duration (seconds) | 1,427 | 778 | 9 | 87,536 |

The wide dispersion in view and like counts reflects the intentional inclusion of channels spanning a large range of audience sizes, from relatively modest channels to major broadcasters with hundreds of millions of views per video. Similarly, the duration range (9 seconds to over 24 hours) captures both short-form content and extended documentary material, providing variability in content type relevant to clickbait detection research. **Figure 6** shows the per-channel breakdown of clickbait and non-clickbait video counts for the 25 most represented channels. **Figure 7** presents distributional characteristics of key content metrics across all 21,238 videos, including title length, description length, view counts, like counts, and a class-level text comparison. **Figure 8** illustrates the distribution of content categories across the 40 channels in the dataset.

The distribution of video content across channels reveals a clear pattern in the prevalence of sensationalized versus straightforward titles. Channels focused on news and documentary content, such as BBC News, Reuters, CNA Insider, National Geographic, and DW Documentary, tend to produce largely factual and informative videos. In contrast, entertainment-oriented channels, including MrBeast, BRIGHT SIDE, and Mo Vlogs, display a higher proportion of attention-grabbing titles. This trend suggests that the use of sensationalized content is strongly associated with channel type and genre, rather than being evenly distributed across the dataset. A visual summary of this distribution is presented in **Figure 6** as a stacked bar chart.

The dataset's content metrics exhibit diverse characteristics, as illustrated in the multi-panel visualization in **Figure 7**. The top-left pie chart shows the overall label distribution, with 54.5% of the videos labeled as non-clickbait and 45.5% as clickbait, indicating a relatively balanced dataset. The top-center histogram indicates that video title lengths are approximately normally distributed, with a mean of 56.7 characters and a median of 55.0, suggesting that most titles are of moderate length. In contrast, the top-right histogram reveals that video description lengths are right-skewed, with a mean of 1,173.7 characters and a median of 932.0, indicating that while many videos have concise descriptions, a subset contains substantially longer text. The bottom-left histogram shows view counts on a log10 scale, with a mean of 6,057,442 and a median of 344,256, reflecting a multimodal distribution that captures both niche and highly popular content. Similarly, the bottom-center histogram of like counts, also on a log10 scale, follows a broadly bell-shaped distribution with a mean of 149,677 and a median of 5,737, highlighting variability in audience engagement across the dataset.

Together, these visualizations provide a comprehensive overview of content characteristics, demonstrating variability in length, reach, and engagement while capturing the dataset's overall structure.

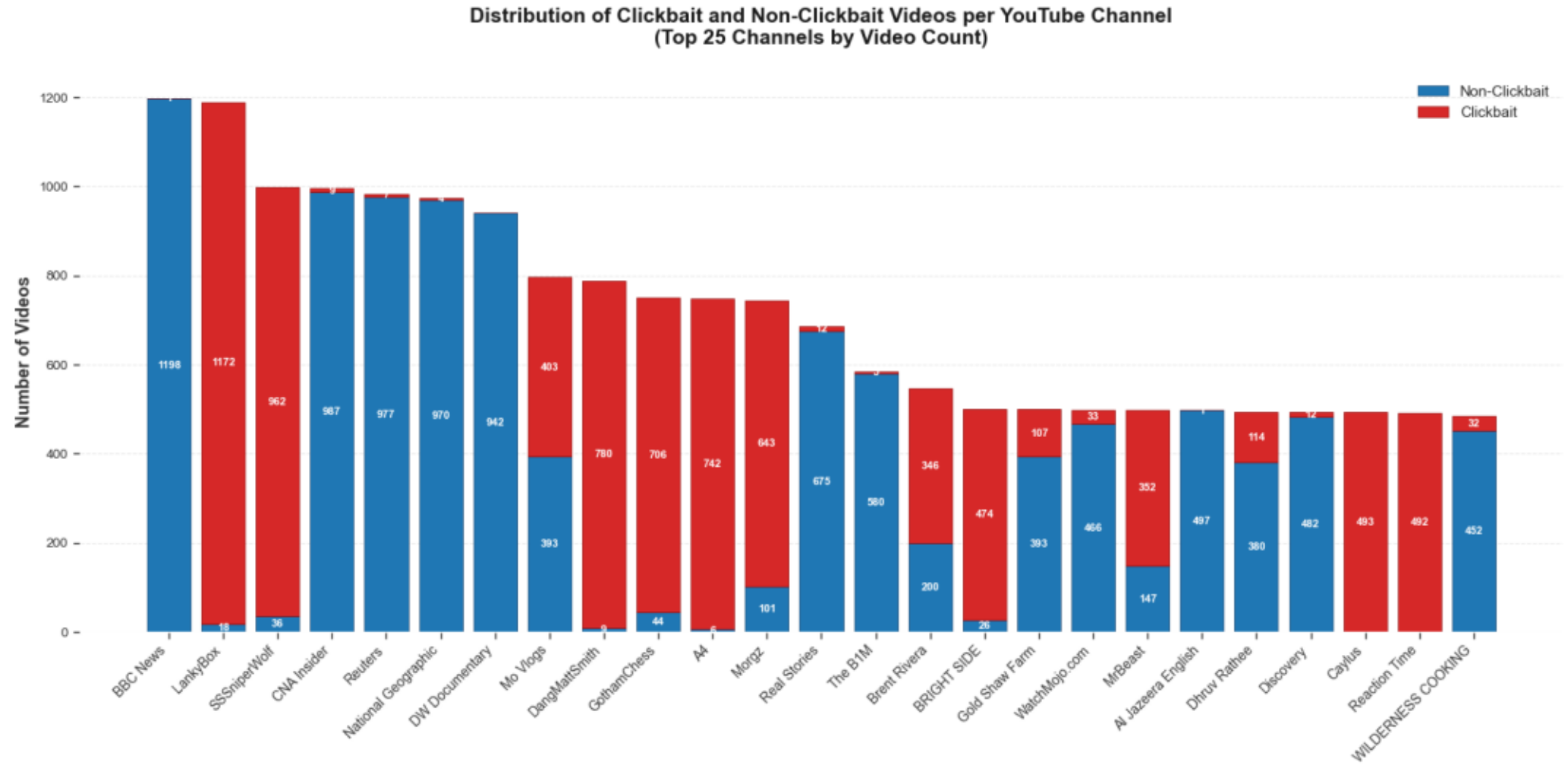

*Figure 6. Distribution of clickbait and non-clickbait videos across the top 25 YouTube channels*

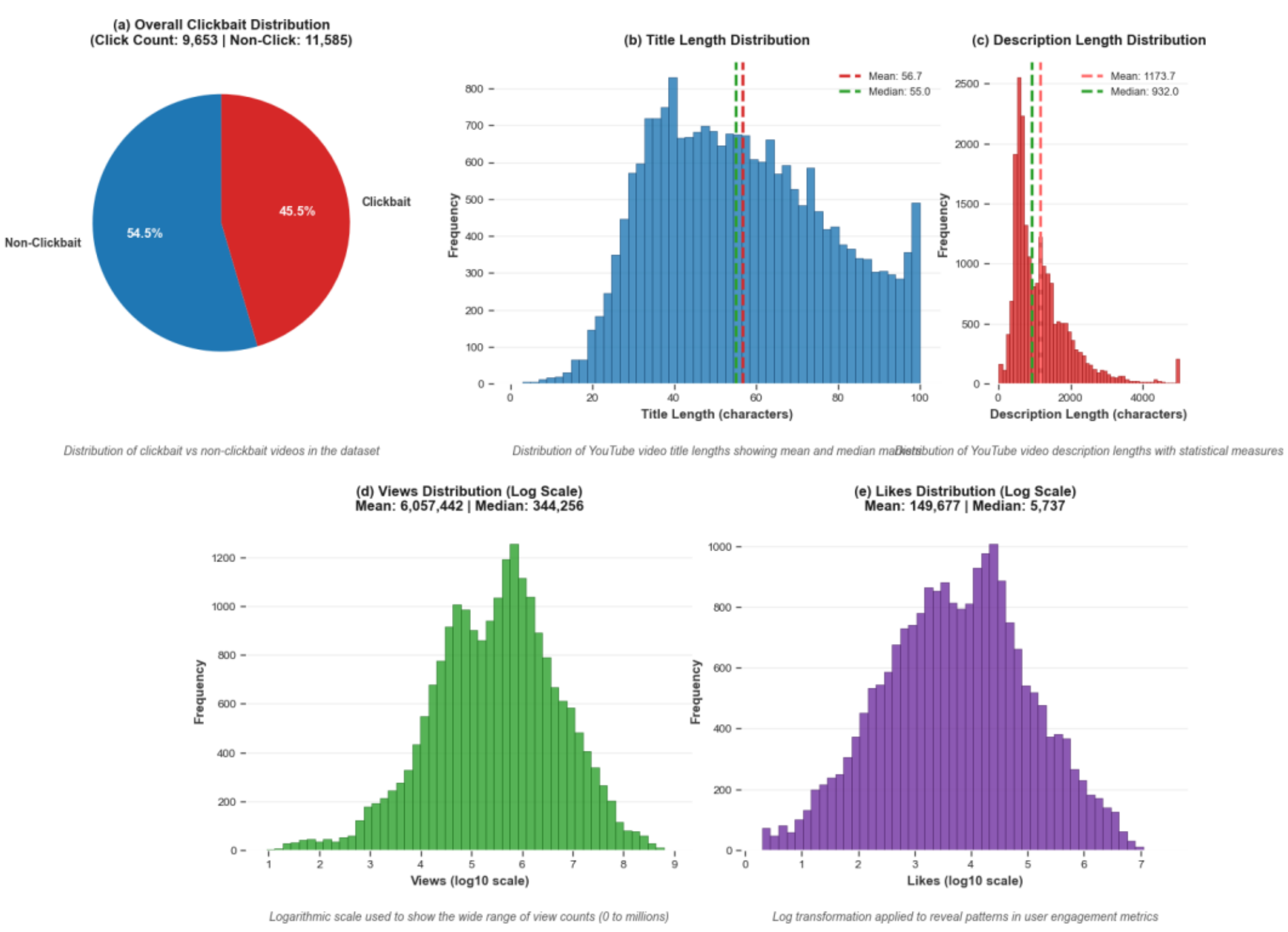

*Figure 7. Distribution of content metrics, including label balance, title length, description length, and engagement statistics*

The dataset encompasses a diverse range of content categories across the 40 channels, as illustrated in **Figure 8**. Entertainment (20.9%) and News (18.6%) are the most prevalent categories, collectively representing nearly 40% of all channels. Education (11.6%) and Gaming (7.0%) form the next tier of representation, while the remaining 16 categories:

including Reaction, Science, Documentary, Religious, Cooking, Kids/Gaming, Lifestyle, Sports, Travel, Comedy, Adventure, Engineering, Farming, ASMR, and Music, each account for between 2.3% and 4.7% of channels. This broad distribution ensures that no single content type dominates the dataset. Such genre diversity is important, as audience expectations and publishing norms vary significantly across categories, which in turn influences how attention-grabbing strategies are employed.

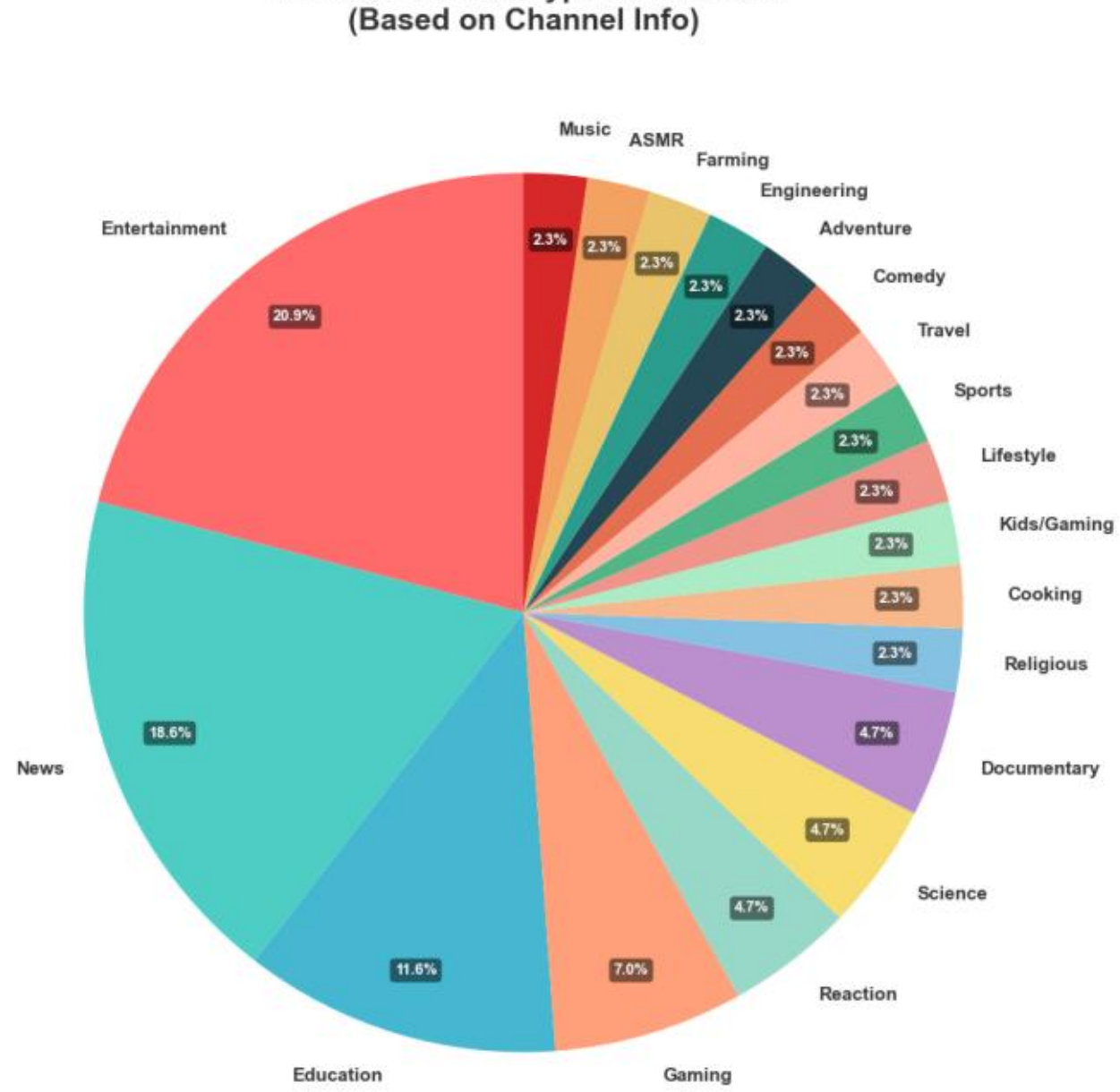

***Figure 8. Distribution of content categories across selected YouTube channels in the dataset***

# Technical Validation

To assess the reliability of the dataset and the annotation process, three quantitative validation analyses were performed: (1) multi-rater agreement using Fleiss' Kappa, (2) a robustness check with Krippendorff's Alpha, and (3) annotator confidence score consistency using pairwise correlation. These analyses evaluate agreement among annotators, robustness to missing annotations, and consistency of annotator certainty. **Table 7** summarizes the scores.

# Multi-Rater Agreement: Fleiss' Kappa

Fleiss' Kappa (κ) is used to measure the level of agreement among multiple annotators while accounting for agreement that may occur by chance. For a dataset with $N$items, $n$annotators, and $k$ categories, it is defined as:

$$\kappa = \frac{\bar{P} - \bar{P}_e}{1 - \bar{P}_e} \quad (1)$$

where the observed agreement is given by:

$$\bar{P} = \frac{1}{N}\sum_{i=1}^{N}\frac{1}{n(n-1)}\sum_{j=1}^{k} n_{ij}\,(n_{ij} - 1) \quad (2)$$

and the expected agreement by chance is:

$$\bar{P}_e = \sum_{j=1}^{k} \left( \frac{1}{N} \sum_{i=1}^{N} n_{ij} \right)^2 \tag{3}$$

When applied to this dataset, Fleiss' Kappa yields a value of $\kappa = 0.6493$, which indicates **substantial agreement** according to the Landis and Koch scale. This result suggests that the annotations are reasonably consistent across all three raters.

# Robustness Check: Krippendorff's Alpha

Krippendorff's Alpha ($\alpha$) provides a reliability measure that accommodates missing annotations. For nominal data, it is defined as:

$$\alpha = 1 - \frac{D_o}{D_e} \tag{4}$$

In this dataset, all annotation records were complete, with no missing labels across annotators. Krippendorff's Alpha was computed at the nominal level, yielding $\alpha = 0.6493$, which closely matches Fleiss' Kappa ($\Delta < 0.001$). This strong alignment indicates consistent reliability across different agreement measures and confirms the robustness of the annotation process.

# Confidence Score Consistency

To assess consistency in annotator certainty, pairwise Pearson correlation coefficients were computed between confidence scores. For two annotators $X$and $Y$, the correlation is:

$$r_{XY} = \frac{\sum_{i=1}^{n} (X_i - \bar{X})(Y_i - \bar{Y})}{\sqrt{\sum_{i=1}^{n} (X_i - \bar{X})^2 \sum_{i=1}^{n} (Y_i - \bar{Y})^2}} \tag{5}$$

The correlation between Annotator 1 and Annotator 2 was relatively low but statistically significant ($r = 0.2393$, $p < 0.001$), with similarly weak yet significant correlations observed across other annotator pairs, resulting in a mean confidence correlation of $r = 0.1546$. This indicates that while annotators consistently assigned high confidence scores, the degree of certainty diverse across individuals, reflecting differences in subjective judgment rather than inconsistency. Overall, the mean confidence score was 4.79 out of 5.00, with the majority of annotations receiving high confidence values ($\geq 4.5$), demonstrating strong certainty in labeling decisions. Although the mean confidence score across annotators was high (4.79 out of 5), the correlation between annotators' confidence scores was relatively low. This suggests that while annotators generally agreed on the final labels, they differed in how they interpreted and used the confidence scale.

Such variation is common in subjective annotation tasks, where individual annotators may adopt different calibration strategies (consistently assigning high confidence versus using a broader range of values). Importantly, this does not undermine label reliability, as reflected by the substantial agreement scores, but rather highlights differences in perceived certainty.

*Table 7. Summary of technical validation metrics assessing annotation reliability and confidence consistency.*

| Validation Test | Statistic | Description |
|---|---|---|
| Fleiss' Kappa | κ = 0.6493 | Substantial agreement among all three annotators; confirms pairwise Cohen's Kappa estimates and indicates strong overall labeling consistency. |
| Krippendorff's Alpha | α = 0.6493 | Provides a robustness check accounting for annotation variability; closely aligns with Fleiss' Kappa, supporting the stability of agreement despite methodological differences. |
| Confidence Correlation | r = 0.1546 ($p < 0.001$) | Low but significant correlation across annotators' confidence scores, indicating variability in perceived certainty while maintaining overall high confidence (mean = 4.79/5.00). |

To assess dataset longevity and accessibility, all YouTube URLs were verified at the time of submission. All videos remained publicly accessible. This high availability rate indicates that the dataset retains strong usability for future research. Users can retrieve video content and metadata dynamically using the provided URLs and the YouTube Data API.

# Usage Notes

The dataset is provided as a UTF-8 encoded CSV file and can be directly loaded in standard analysis environments such as Python, R, or spreadsheet software. The final_label column is recommended for binary clickbait classification, while individual annotator labels and confidence scores enable soft-label training, confidence-weighted learning, or disagreement analysis. Text fields (title and description) support natural language processing tasks, while thumbnail images, provided as local files in JPG and WebP formats via the local_thumbnail field, support visual and multimodal analysis. Video content and audio transcriptions are not packaged with the dataset due to storage constraints; researchers who require these modalities can retrieve them directly from YouTube using the URL field provided for each record. Users should apply standard preprocessing and null-handling as appropriate for their model pipelines.

# Funding

Not Applicable

# Data Availability

The dataset is publicly available on Figshare at https://doi.org/10.6084/m9.figshare.31891576 under the CC BY 4.0 license. The deposit includes Final_data.csv (33.25 MB, 21,238 videos) with metadata, annotations, and final labels, plus a folder of thumbnail images (2.41 GB). Video and audio content can be retrieved via the provided YouTube URLs.

# Code Availability

All custom code for data collection, annotation, label aggregation, and analysis is publicly available at GitHub repository (https://github.com/minhazulruet/Clickbait-data-collect-and-label). The repository includes the FastAPI-based scraping scripts, the PHP/MySQL annotation platform, and Python scripts/notebooks for data processing and technical validation. A requirements.txt file is provided to reproduce the analysis environment.

# Acknowledgements

The authors would like to express their gratitude to ChatGPT, and Grammarly, among other AI tools, for their invaluable help in polishing and enhancing the writing's quality.

# Author Contributions

M.I.: Conceptualization, Methodology, Formal Analysis, Software, Investigation, Writing – original draft (methodology and coding); T.J.: Investigation, Data Curation, Formal Analysis, Methodology (algorithm design assistance), Visualization, Writing – original draft (literature review); A.K.: Supervision, Software, Resources, Validation, Project administration, Conceptualization, Methodology; S.S.: Data Curation, Formal Analysis, Software, Writing – review & editing; S.R.: Data Curation, Formal Analysis, Investigation, Writing – review & editing; M.M.: Supervision, Methodology, Software, Visualization; M.A.A.: Funding Acquisition, Conceptualization (planning), Validation, Supervision, Writing – review & editing; H.N.: Validation, Writing – review & editing. All authors contributed to reviewing and finalizing the manuscript.

# Competing Interests

The authors declare no competing interests.

# References

1. Chakraborty, A., Paranjape, B., Kakarla, S. & Ganguly, N. Stop Clickbait: Detecting and preventing clickbaits in online news media. *Proceedings of the 2016 IEEE/ACM International Conference on Advances in Social Networks Analysis and Mining, ASONAM 2016* 9–16 (2016) doi:10.1109/ASONAM.2016.7752207.
2. Potthast, M., Gollub, T., Hagen, M. & Stein, B. The Clickbait Challenge 2017: Towards a Regression Model for Clickbait Strength. http://arxiv.org/abs/1812.10847 (2018).
3. Biyani, P., Tsioutsiouliklis, K. & Blackmer, J. '8 Amazing Secrets for Getting More Clicks': Detecting Clickbaits in News Streams Using Article Informality. *Proceedings of the AAAI Conference on Artificial Intelligence* **30**, 94–100 (2016).
4. Potthast, M., Gollub, T., Hagen, M. & Stein, B. The Clickbait Challenge 2017: Towards a Regression Model for Clickbait Strength. https://arxiv.org/pdf/1812.10847 (2018).
5. Al-Sarem, M. *et al.* An Improved Multiple Features and Machine Learning-Based Approach for Detecting Clickbait News on Social Networks. *Applied Sciences 2021, Vol. 11, Page 9487* **11**, 9487 (2021).
6. Bronakowski, M., Al-khassaweneh, M. & Al Bataineh, A. Automatic Detection of Clickbait Headlines Using Semantic Analysis and Machine Learning Techniques. *Applied Sciences 2023, Vol. 13, Page 2456* **13**, 2456 (2023).
7. Liu, T. *et al.* Clickbait detection on WeChat: A deep model integrating semantic and syntactic information. *Knowl. Based. Syst.* **245**, 108605 (2022).
8. Muqadas, A. *et al.* Deep learning and sentence embeddings for detection of clickbait news from online content. *Scientific Reports 2025 15:1* **15**, 13251- (2025).
9. Nguyen, D. P., Tran, T. K., Nguyen, Y. M. & Vo, B. ViClickbait-2025: A comprehensive dataset for Vietnamese clickbait detection. *Data Brief* **63**, 112164 (2025).
10. Elyashar, A., Bendahan, J. & Puzis, R. Detecting Clickbait in Online Social Media: You Won't Believe How We Did It. *Lecture Notes in Computer Science (including subseries Lecture Notes in Artificial Intelligence and Lecture Notes in Bioinformatics)* **13301 LNCS**, 377–387 (2022).
11. Gamage, B., Labib, A., Joomun, A., Lim, C. H. & Wong, K. S. Baitradar: A multi-model clickbait detection algorithm using deep learning. *ICASSP, IEEE International Conference on Acoustics, Speech and Signal Processing - Proceedings* **2021-June**, 2665–2669 (2021).
12. Zannettou, S., Chatzis, S., Papadamou, K. & Sirivianos, M. The Good, the Bad and the Bait Detecting and Characterizing Clickbait on YouTube. in *2018 IEEE Security and Privacy Workshops (SPW)* 63–69 (2018). doi:10.1109SPW.2018.00020.
13. Vitadhani, A., Ramli, K. & Dewi Purnamasari, P. Detection of Clickbait Thumbnails on YouTube Using Tesseract-OCR, Face Recognition, and Text Alteration. *ICAICST 2021 - 2021 International Conference on Artificial Intelligence and Computer Science Technology* 56–61 (2021) doi:10.1109/ICAICST53116.2021.9497811.

14. Varshney, D. & Vishwakarma, D. K. A unified approach for detection of Clickbait videos on YouTube using cognitive evidences. *Applied Intelligence 2021 51:7* **51**, 4214–4235 (2021).
15. Mowar, P., Jain, M., Goel, R. & Vishwakarma, D. K. Clickbait in YouTube Prevention, Detection and Analysis of the Bait using Ensemble Learning. http://arxiv.org/abs/2112.08611 (2021).
16. Sung, Y. Y. & Boyd-Graber, J. Not all Fake News is Written: A Dataset and Analysis of Misleading Video Headlines. *EMNLP 2023 - 2023 Conference on Empirical Methods in Natural Language Processing, Proceedings* 16241–16258 (2023) doi:10.18653/v1/2023.emnlp-main.1010.
17. Rahman, S. S., Das, A., Sharif, O. & Hoque, M. M. Identification of Deceptive Clickbait Youtube Videos Using Multimodal Features. *Lecture Notes in Networks and Systems* **853 LNNS**, 199–208 (2023).
18. Al-Ali, M. N. & Hamzeh, M. S. M. Extra cues extra views: A multimodal detection of Arabic clickbait thumbnail verbo-visual cues. *Discourse and Communication* **18**, 3–27 (2024).
19. Imran, A. Al, Shovon, M. S. H. & Mridha, M. F. BaitBuster-Bangla: A comprehensive dataset for clickbait detection in Bangla with multi-feature and multi-modal analysis. *Data Brief* **53**, 110239 (2024).
20. Naveed, W., Uzmi, Z. A. & Qazi, Z. A. ThumbnailTruth: A Multi-Modal LLM Approach for Detecting Misleading YouTube Thumbnails Across Diverse Cultural Settings. http://arxiv.org/abs/2509.04714 (2025).
21. Kim, J., Shin, Y., Jung, G. & Ahn, H. A hybrid detection method for YouTube fake news using related video data. *Eng. Appl. Artif. Intell.* **156**, 111130 (2025).
22. Zhang, Y. *et al.* Multimodal graph contrastive learning for fake news video detection. *Journal of King Saud University Computer and Information Sciences 2025 38:1* **38**, 11- (2025).
23. Zheng, J., Yu, K. & Wu, X. A deep model based on Lure and Similarity for Adaptive Clickbait Detection. *Knowl. Based. Syst.* **214**, 106714 (2021).
24. Goodfellow, I., Bengio, Y., Courville, A. & Heaton, J. Ian Goodfellow, Yoshua Bengio, and Aaron Courville: Deep learning. *Genetic Programming and Evolvable Machines 2017 19:1* **19**, 305–307 (2017).
25. Varshney, D. & Vishwakarma, D. K. A unified approach for detection of Clickbait videos on YouTube using cognitive evidences. *Applied Intelligence 2021 51:7* **51**, 4214–4235 (2021).